\documentclass{article} 
\usepackage[dvipsnames]{xcolor}
\usepackage{iclr2021_conference,times}

\usepackage{amsmath,amsfonts,bm}

\def\eqref#1{equation~\ref{#1}}

\def\1{\bm{1}}

\DeclareMathAlphabet{\mathsfit}{\encodingdefault}{\sfdefault}{m}{sl}
\SetMathAlphabet{\mathsfit}{bold}{\encodingdefault}{\sfdefault}{bx}{n}

\usepackage{url}
\usepackage{booktabs} 
\usepackage{graphicx}
\usepackage{rotating}
\usepackage{tabu}
\usepackage{wrapfig}
\usepackage{caption}
\usepackage{subcaption}
\usepackage{amssymb}
\usepackage{pifont}

\usepackage[colorinlistoftodos,textsize=tiny, textwidth=18mm]{todonotes}

\title{How Sensitive are Meta-Learners\\to Dataset Imbalance?}

\author{
Mateusz Ochal$^{1,2}$,~Massimiliano Patacchiola$^2$,~Amos Storkey$^2$,~Jose Vazquez$^{3}$,~Sen Wang$^1$ \\
$^1$School of Engineering and Physical Sciences, Heriot-Watt University, UK \\
$^2$School of Informatics, University of Edinburgh, UK \\
$^3$SeeByte Ltd., Edinburgh, UK \\
\texttt{m.ochal@hw.ac.uk, mpatacch@ed.ac.uk, a.storkey@ed.ac.uk} \\
\texttt{jose.vazquez@seebyte.com, s.wang@hw.ac.uk}
}

\newif\ifincludecomment
\includecommenttrue 

\ifincludecomment
  
\else
  \NewEnviron{guidance}{}{}
\fi
\ifincludecomment
\newcommand{\maybecomment}[1]{\todo[color=olive!40]{#1}} 
\newcommand{\maybetohere}[1]{\todo[color=red!40]{#1}} 
\newcommand{\maybedelete}[1]{\todo[color=blue!40]{#1}} 

\else
  \newcommand{\maybecomment}[1]{}
  
\newcommand{\maybedelete}[1]{} 
\fi
\newcommand{\amostohere}[1]{{\color{black}\maybetohere{AMOS HERE}}}

\newcommand\longtail{long$-$tail}

\usepackage{array}
\newcolumntype{H}{>{\setbox0=\hbox\bgroup}c<{\egroup}@{}}

\makeatletter
\newcommand{\ssymbol}[1]{$\ ^{\@fnsymbol{#1}}$}
\makeatother
\newcommand{\specialcell}[2][c]{%
  \begin{tabular}[#1]{@{}c@{}}#2\end{tabular}}

\iclrfinalcopy 
\begin{document}

\maketitle

\begin{abstract}
    Meta-Learning (ML) has proven to be a useful tool for training Few-Shot Learning (FSL) algorithms by exposure to batches of tasks sampled from a meta-dataset. However, the standard training procedure overlooks the dynamic nature of the real-world where object classes are likely to occur at different frequencies. While it is generally understood that imbalanced tasks harm the performance of supervised methods, there is no significant research examining the impact of imbalanced meta-datasets on the FSL evaluation task. This study exposes the magnitude and extent of this problem. Our results show that ML methods are more robust against meta-dataset imbalance than imbalance at the task-level with a similar imbalance ratio ($\rho<20$), with the effect holding even in long-tail datasets under a larger imbalance ($\rho=65$). Overall, these results highlight an implicit strength of ML algorithms, capable of learning generalizable features under dataset imbalance and domain-shift. The code to reproduce the experiments is released under an open-source license\footnote{ \url{https://github.com/mattochal/imbalanced_fsl_public}}.
\end{abstract}
 
 \section{Introduction}
 Few-Shot Learning (FSL) aims at reducing the burden of training machine-learning models on a large number of labeled data points. A common way of training FSL models is through Meta-Learning (ML) \citep{Hospedales2020,Vinyals2017matching} with the model repeatedly exposed to batches of FSL tasks/episodes sampled from a (meta-)training dataset that is different but similar to the one seen during the (meta-)testing phase. We will use the prefix \emph{“meta”} to distinguish the high-level training and evaluation routines of ML (outer loop) from the training and evaluation routines at the single-task level (inner loop).
 
\paragraph{Motivation.} Standard FSL benchmarks (eg. Omniglot, Mini-ImageNet) assume an equal number of data points for each class in the meta-dataset. However, in real-world applications, it is common to encounter datasets with varying numbers of data points per class \citep{Guan2020aerial,Ochal2020oceans,massiceti2021orbit}. In some cases, the dataset distribution can be immeasurably large or even unknown, like in online class imbalance \citep{wang2014resampling}. Although meta-dataset imbalance is present in recent FSL benchmarks, such as Meta-Dataset \citep{Triantafillou2019meta} or meta-iNat \citep{Wertheimer2019metainat}, its impact on ML has remained unexplored.

\paragraph{Contributions.} We provide quantitative insights into the dataset imbalance problem and show that meta-learners are robust to meta-dataset imbalance under 1) various imbalance distributions, 2) dataset sizes, and 3) moderate cross-domain-shift. The impact of imbalance at the task level is much more significant in comparison. 

 \section{Related Work}\label{sec related}
 \paragraph{Class Imbalance.} In classification, imbalance occurs when at least one class (the majority class) contains a higher number of samples than the others. The classes with the lowest number of samples are called minority classes. If uncorrected, conventional supervised loss functions, such as (multi-class) cross-entropy, skew the learning process in favor of the majority class, introducing bias and poor generalization toward the minority class samples \citep{Buda2018imbalance,Leevy2018bigdata}. The object recognition community studies class imbalance using real-world datasets or distributions that approximate real-world imbalance \citep{Buda2018imbalance,Johnson2019imbalance,Liu2019tailed}. \cite{Buda2018imbalance} state that two distributions can be used: \emph{linear} and \emph{step} imbalance. At large-scale, datasets with many samples and classes tend to follow a \emph{long-tail} distribution \citep{Liu2019tailed,Salakhutdinov2011longtail,Reed2001powerlaw}, with most of the classes occurring with small frequency and a few classes occurring with high frequency. 
 
 \paragraph{Meta-Dataset FSL Benchmarks.} FSL methods \citep{Snell2017proto,Sung2017relationnet,Edwards2017,Vinyals2017matching,Finn2017maml,Ravi2017,Chen2019closer,Dhillon2020baseline,Patacchiola2019gpshot,zhang2021shallow} are typically compared against balanced benchmark (eg. Omniglot, MiniImageNet). More recent benchmarks \citep{Triantafillou2019meta, Wertheimer2019metainat} contain some levels of imbalance in the meta-dataset. Specifically, Meta-Dataset \citep{Triantafillou2019meta} combines datasets of different sizes (e.g., Omniglot, CUB, Fungi, Aircraft, ImageNet, etc.) into a single corpus. A different dataset, meta-iNat \citep{Wertheimer2019metainat}, models imbalance according to a long-tail distribution. While imbalance at the \emph{task-level} has received some attention \citep{Triantafillou2019meta, Chen2020mamlstop, Lee2020baysiantaml, Guan2020aerial}, limited research quantifies the impact of imbalance at the \emph{dataset-level}. It can be cumbersome to evaluate the imbalance of Meta-Dataset and meta-iNat specifically, as it requires access to a balanced version of the datasets with an equal total number of samples. Therefore, in our analysis, we artificially induce imbalance into datasets and model imbalance according to various distributions approximating many real-world scenarios. 
 

 \section{Problem Definition}\label{sec methodology}
 
 \paragraph{Standard Task/Meta-Training.} Benchmarking FSL methods typically involves three phases: meta-training/pre-training, meta-validation, and meta-testing. Each phase samples batches of data points or tasks from a separate dataset: $\mathcal{D}_{train}$, $\mathcal{D}_{val}$, and $\mathcal{D}_{test}$, respectively, such that the classes and samples between each of the datasets are non-overlapping. We assume that a dataset $\mathcal{D}$ is $balanced$ when it contains $N^\mathcal{D}$ classes and $K^\mathcal{D}$ samples for each class. Similarly, a standard $K^\mathcal{S}$-shot $N^\mathcal{S}$-way FSL classification task is defined by a small \emph{support set}, $\mathcal{S} = \left\{ (x_1, y_1), \dots , (x_{s}, y_{s}) \right\} \sim \mathcal{D}$, containing $N^\mathcal{S} \times K^\mathcal{S}$ image-label pairs drawn from $N^\mathcal{S}$ unique classes with $K^\mathcal{S}$ samples per class ($|\mathcal{S}|=K^\mathcal{S} \times N^\mathcal{S}$ and $K^\mathcal{S} \ll K^{\mathcal{D}}$ and $N^\mathcal{S} \ll N^{\mathcal{D}}$). The goal is to minimize some loss over a \emph{query set}, $\mathcal{Q} = \left\{ (x_1, y_1), \dots , (x_{q}, y_{q}) \right\}  \sim \mathcal{D}$, containing a different set of $K^\mathcal{Q}$ samples drawn from the same $N^\mathcal{S}$ classes (i.e. $N^\mathcal{Q}=N^\mathcal{S}$, $\mathcal{Q}^{(x)} \cap \mathcal{S}^{(x)} = \emptyset$ and $\mathcal{Q}^{(y)}\equiv \mathcal{S}^{(y)}$).

\paragraph{Imbalanced Tasks/Meta-Dataset.} For brevity, but without loss of generality, we define a distribution for a set of data points $\mathcal{*}\in\{\mathcal{D},\mathcal{S},\mathcal{Q}\}$ as a tuple ($K_{min}^{\mathcal{*}}$, $K_{max}^{\mathcal{*}}$, $N^{\mathcal{*}}$, $M^{\mathcal{*}}$) for a distribution $\mathcal{I} \in \{linear,step,\longtail\}$ \citep{Buda2018imbalance,Wertheimer2019metainat}, where $K_{min}^{\mathcal{*}}$ is the minimum number of samples per class, $K_{max}^{\mathcal{*}}$ is the maximum number of samples per class, $N^{\mathcal{*}}$ is the number of classes, and $M^{\mathcal{*}}$ is an additional parameter used for $step$ and $\longtail$ imbalance. We induce imbalance in our experiments using one of $I$ distributions defined as:
 \begin{itemize}
     \item \emph{Linear imbalance}. The $K^\mathcal{*}_{i}$ number of samples for each class $i\in\{1..N^\mathcal{*}\}$ is defined by: 
     \begin{equation}
        \begin{split}
            K^\mathcal{*}_{i} = \mathtt{round} (
            K^\mathcal{*}_{min}-c + (i - 1) \times (K^\mathcal{*}_{max}+ 2 \times c - K^\mathcal{*}_{min}) / (N^\mathcal{*}-1) ) ,
        \end{split}
     \end{equation}
     where $c=0.499$ for rounding purposes. For example, this means that for $linear$ (1,9,5,-) set,  $K^\mathcal{*}_{i} \in \{1, 3, 5, 7, 9\}$, and for $linear$ (4,6,5,-) set, $K^\mathcal{*}_{i} \in \{4, 4, 5, 6, 6\}$. 
     \item \emph{Step imbalance}. The number of class samples, $K^\mathcal{*}_{i}$, is determined by an additional variable $M^\mathcal{*}$ specifying the number of minority classes. Specifically, for classes $i\in\{1..N^\mathcal{*}\}$:
     \begin{equation}
         K^\mathcal{*}_{i}=
         \begin{cases}
             K^\mathcal{*}_{min},           &    \text{if}~ i \leq M^\mathcal{*},\\
             K^\mathcal{*}_{max},           &    \text{otherwise}.
         \end{cases}
    \end{equation}
     For example, in a $step$ (1,9,5,1) set, there is 1 minority class, and $K^\mathcal{*}_{i} \in \{1, 9, 9, 9, 9\}$. 
     \item \emph{Long-Tail imbalance}. Imbalance could be modeled by a Zipf's/Power Law \citep{Reed2001powerlaw} for a more realistic imbalance distribution.
 \end{itemize}
  We also report the imbalance ratio $\rho$, which is a scalar identifying the level of class imbalance; this is often reported in the CI literature for the supervised case \citep{Buda2018imbalance}. We define $\rho$ to be the ratio between the number of samples in the majority and minority classes in a set of data points:
 \begin{equation}
   \rho=\frac{K^\mathcal{*}_{max}}{K^\mathcal{*}_{min}}.
 \end{equation}
 
\section{Experiments}\label{sec exp}

\paragraph{Setup.} We meta-trained a range of FSL and ML methods on multiple datasets and imbalance distributions. Specifically, we investigate Prototypical Networks \citep{Snell2017proto}, Matching Networks \citep{Vinyals2017matching}, Relation Networks \citep{Sung2017relationnet}, MAML \citep{Finn2017maml}, ProtoMAML \citep{Triantafillou2019meta}, DKT \citep{Patacchiola2019gpshot}, SimpleShot \citep{Wang2019simpleshot}, and supervised pre-training methods -- Baseline and Baseline++ from \cite{Chen2019closer}.
Each model was meta-trained/pre-trained on 100k tasks/mini-batches sampled from variations of the meta-training dataset of Mini-ImageNet \citep{Ravi2017} ($\mathcal{D}_{train}$), originally containing 64 classes and 600 samples per class. To get enough samples for the majority classes, we halved the total number of samples from the original $\mathcal{D}_{train}$ of Mini-ImageNet, and denote this dataset as $\mathcal{D}'_{train}$, where $|\mathcal{D}'_{train}| \approx 64 \times 300 = 19200$. We induced imbalance into the dataset according to one of the $\mathcal{I}$-distributions outlined in Section~\ref{sec methodology}. For simplicity, we did not modify the meta-validation and meta-testing datasets ($\mathcal{D}_{val}$ and $\mathcal{D}_{test}$), keeping them like in the original Min-ImageNet. To emulate stronger domain-shift, we also evaluated models on 50 randomly selected classes from CUB-2011 \citep{Wah2011cub}. More details and experiments are in Appendix~\ref{sec: implement}.

    

\begin{figure}
    \centering
    \includegraphics[width=0.99\textwidth]{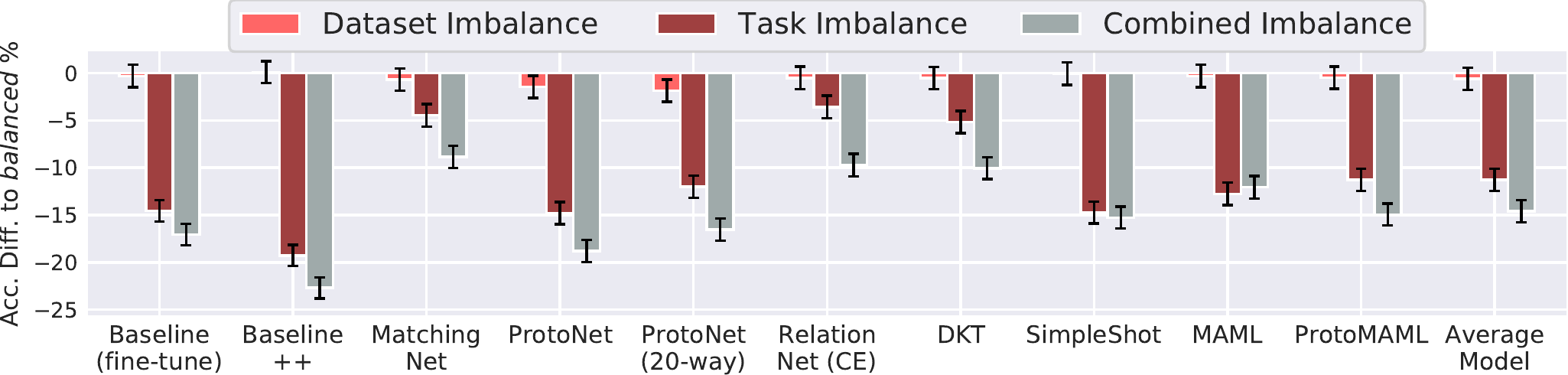}
    \caption{Difference in accuracy between dataset imbalance (light red), task imbalance (dark red), and combined imbalance (grey) against the $balanced$ meta-training baseline. Negative performance indicates a lower accuracy compared to $balanced$. The results suggest that models are quite robust to dataset imbalance compared to imbalance at the task-level.}
    \label{fig linear_task_vs_dataset}
\end{figure}

\paragraph{Meta-learners are robust to dataset imbalance.} We compare dataset imbalance to task imbalance using a similar distribution. In the \emph{dataset imbalance} condition, tasks (5-shot, 5-way) are sampled from a a linearly-imbalanced dataset $\mathcal{D}'_{train}$ (30,570,64,-). In the \emph{task imbalance} condition, tasks are sampled from a linearly-imbalanced distribution (1,9,5,-) -- also called 1-9shot 5-way -- and from a balanced dataset $\mathcal{D}'_{train}$ (300,300,64,-). Finally, in the \emph{combined imbalance} condition imbalanced tasks (1,9,5,-) are sampled from the imbalanced dataset $\mathcal{D}'_{train}$ (30,570,64,-).  Figure~\ref{fig linear_task_vs_dataset} shows the difference in $\mathcal{D}_{test}$ accuracy between dataset imbalance (light red) and task imbalance (dark red) compared against the score of standard training ($balanced$ at both dataset and task levels). Dataset imbalance causes an insignificant drop in performance, being mostly within error bars with respect to $balanced$. On the other hand, the imbalanced task condition causes a significant drop in performance, up to $-20\%$ for some methods despite the slightly smaller imbalance magnitude ($\rho=9$ c.f. $\rho=19$). This result suggests that meta-learners are quite robust to dataset imbalance.  Interestingly, the combined task and dataset imbalance has a slight compounding effect on performance, yielding $-15\%$ on average and performing worse than the combined sum of the two imbalance conditions taken individually ($-12\%$). Simpleshot and MAML are the only methods where the difference between combined imbalance and task imbalance is not significant, suggesting that they could be particularly resistant to this compound effect.

\begin{table}[t]
    \caption{Evaluation accuracy after meta-training on $\mathcal{D}'_{train}$ derived from Mini-ImageNet with various imbalance distributions. Small differences in accuracy between $balanced$ and other distributions, suggest a small effect of imbalance at dataset level. \emph{Left:} Evaluation on the meta-testing dataset of Mini-ImageNet. \emph{Right:} Evaluation on the meta-testing dataset of CUB.
    } \label{tbl imbalanced_meta_dataset_5shot_5way}
    \centering
    \scalebox{0.68}{ \begin{tabular}{r|cHccc|cHccc}
\toprule
$\mathcal{D}'_{train} \rightarrow \mathcal{D}_{test}$ & \multicolumn{5}{c|}{Mini-ImageNet $\rightarrow$ Mini-ImageNet} & \multicolumn{5}{c}{Mini-ImageNet $\rightarrow$ CUB} \\
Imbalance in $\mathcal{D}'_{train}$  &  \specialcell{ $balanced$ \\ {\scriptsize(300, 300, 64, -) }} &  \specialcell{ $linear$ \\ {\scriptsize (30, 570, 64, -)} } & \specialcell{  $step$-32 \\ {\scriptsize(30, 570, 64, 32)} } & \specialcell{  $step$-22 \\ {\scriptsize(25, 444, 64, 22)} } & \specialcell{  $step$-animal \\ {\scriptsize(25, 444, 64, 22)} }  &  \specialcell{ $balanced$ \\ {\scriptsize(300, 300, 64, -) }} &  \specialcell{ $linear$ \\ {\scriptsize (30, 570, 64, -)} }  & \specialcell{  $step$-32 \\ {\scriptsize(30, 570, 64, 32)} } & \specialcell{  $step$-22 \\ {\scriptsize(25, 444, 64, 22)} }  & \specialcell{  $step$-animal \\ {\scriptsize(25, 444, 64, 22)} } \\
\midrule
Baseline (fine-tune) &           61.23\textcolor{gray}{\tiny{$\pm$0.72}} &           60.93\textcolor{gray}{\tiny{$\pm$0.70}} &           60.44\textcolor{gray}{\tiny{$\pm$0.71}} &           60.36\textcolor{gray}{\tiny{$\pm$0.71}} &           59.32\textcolor{gray}{\tiny{$\pm$0.73}} &           56.61\textcolor{gray}{\tiny{$\pm$0.72}} &           56.43\textcolor{gray}{\tiny{$\pm$0.73}} &           55.83\textcolor{gray}{\tiny{$\pm$0.71}} &           56.35\textcolor{gray}{\tiny{$\pm$0.72}} &           52.83\textcolor{gray}{\tiny{$\pm$0.73}} \\
Baseline++           &           63.86\textcolor{gray}{\tiny{$\pm$0.67}} &           63.97\textcolor{gray}{\tiny{$\pm$0.67}} &  \textbf{63.23\textcolor{gray}{\tiny{$\pm$0.66}}} &  \textbf{62.91\textcolor{gray}{\tiny{$\pm$0.67}}} &           62.03\textcolor{gray}{\tiny{$\pm$0.70}} &           57.71\textcolor{gray}{\tiny{$\pm$0.72}} &           57.33\textcolor{gray}{\tiny{$\pm$0.74}} &           56.62\textcolor{gray}{\tiny{$\pm$0.75}} &  \textbf{57.07\textcolor{gray}{\tiny{$\pm$0.75}}} &           52.07\textcolor{gray}{\tiny{$\pm$0.71}} \\
\bottomrule
\end{tabular}  } 
\end{table}

\paragraph{Robustness against domain-shift.} For the remaining sections, we kept the tasks balanced (5-shot 5-way), digging deeper into dataset imbalance. In this paragraph, we examined various $step$ imbalance distributions and evaluated performance with domain-shift. Table~\ref{tbl imbalanced_meta_dataset_5shot_5way} shows the meta-testing\begin{wrapfigure}{r}{0.4\textwidth}
  \begin{center}
    \includegraphics[width=\linewidth]{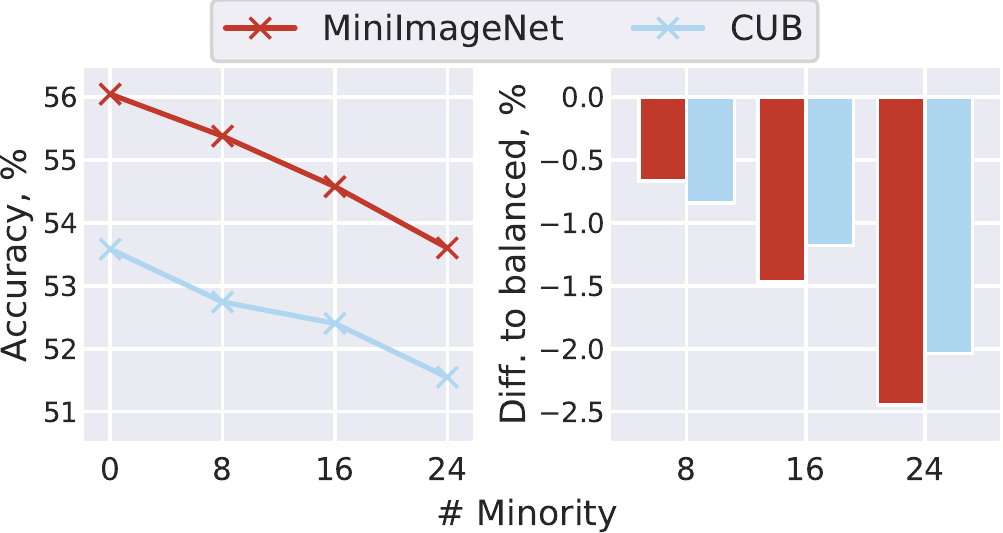}
  \end{center}
  \caption{Combined model average performance with increasing minority classes. \emph{Left:} Combined accuracy of all models. \emph{Right:} Performance difference to the balanced dataset.}\label{fig reduced_meta}
  \vspace{-0.05cm}
\end{wrapfigure} performance on $\mathcal{D}_{test}$ of Mini-ImageNet (left column set) or CUB (right column set) after training on $step$ imbalanced $\mathcal{D}'_{train}$. The bottom row shows the average model difference in accuracy between $balanced$ and the imbalanced datasets. The results show a small negative difference ($-3.1\%$) between the $balanced$ dataset and the step scenario with 32 minority classes ($step$-32). Evaluation of models trained under the $step$-32 distribution on CUB, suggests only a small difference ($-1.6$) under the domain-shift condition. Note that $D'_{train}$ contains 22 animal classes, including 3 classes of birds, and this might not represent a very large domain-shift. In Figure~\ref{fig reduced_meta} and Appendix~\ref{appendix reduced}, we examine more $step$ imbalance distributions on a smaller dataset containing only 1/8\textsuperscript{th} of total samples of $\mathcal{D}_{train}$ from Mini-ImageNet. Overall, the performance drops as the number of minority classes increases, but the effect still remains quite low ($<3\%$ absolute difference).

\paragraph{Robustness against larger domain-shift.} Next, we examined a stronger domain-shift scenario. We set all 22 animal classes in $\mathcal{D}'_{train}$ to contain $K_{min}^{\mathcal{D}}=25$ samples each, and we set the other 42 classes to have $K_{max}^{\mathcal{D}}=444$ samples each. We call this setting ``$step$-animal'' because it reduces the diversity of animal samples seen during meta-training. As a control variable, we examine a similar $step$ imbalance with 22 minority classes picked uniformly at random ($step$-22). Results in Table~\ref{tbl imbalanced_meta_dataset_5shot_5way} suggest that $step$-animal performs slightly worse ($1.6\%$) compared to $step$-22 on average. SimpleShot and the two Baselines are particularly affected by the larger domain-shift on CUB, getting $-4.0\%$ performance drop on $step$-animal compared to $step$-22. Perhaps, this drop is due to the particular training procedure used for these methods, which are pre-trained on mini-batches instead of tasks. This suggests an implicit strength of ML algorithms against larger domain-shift. 

\paragraph{Robustness against larger imbalance.} In Appendix~\ref{appendix longtail}, we examined $\longtail$ distributions with an imbalance ratio $\rho = 65$ on a larger dataset ($6.5\times$ larger than $\mathcal{D}'_{train}$, and derived directly from ImageNet). We observed a more significant drop in accuracy w.r.t. the balanced dataset for ProtoNet and MAML ($-6.3\%$ and $-4.1\%$, respectively). This suggests that a more natural imbalance distribution and a higher imbalance ratio could cause a larger performance drop. However, the performance drop still remained smaller compared to task-level imbalance in Figure~\ref{fig linear_task_vs_dataset}.

\section{Discussion and Conclusion}\label{sec conclude}
In this work, we have provided insights into the meta-dataset imbalance problem in meta-learning, showing that models are quite robust to meta-dataset imbalance -- meaning that they would likely experience only small drops in accuracy points when exposed to dataset imbalance in the real-world. In contrast, the support-set imbalance yields a significantly larger (an order of magnitude) performance drop.
Overall, our results suggest that dataset imbalance has a small negative effect on the ML procedure when tasks are balanced. This seems to point to an implicit strength of ML algorithms, which can learn generalizable features even when exposed to imbalanced meta-datesets. Higher levels of step and long-tail imbalance will likely produce more dramatic performance differences. In this study, we evaluated generalization to novel tasks and classes only - however, it is likely that imbalance in $\mathcal{D}_{train}$ would impact performance on base tasks and classes sampled from $D_{train}$. Moreover, in this study, we were constrained by the size of meta-training dataset of Mini-ImageNet and ImageNet as we had to control the size of the dataset. In the real-world, meta-datasets can be very large with very high imbalance ratios ($\rho\gg100$) and following long-tail distributions \citep{Liu2019tailed,Salakhutdinov2011longtail}. Future work should investigate how these larger imbalance levels could affect meta-learning and cross-domain generalization.

\section*{Acknowledgment}
We want to thank Eleni Triantafillou, Hae Beom Lee, Hayeon Lee, and the members of the Bayesian and Neural Systems group at the University of Edinburgh for valuable comments, suggestions, and discussions offered at various stages of this work. This work was supported by the EPSRC Centre for Doctoral Training in Robotics and Autonomous Systems, funded by the UK Engineering and Physical Sciences Research Council (Grant No. EP/S515061/1) and SeeByte Ltd, Edinburgh, UK.

\bibliography{bib/bib}

\begin{thebibliography}{33}
\providecommand{\natexlab}[1]{#1}
\providecommand{\url}[1]{\texttt{#1}}
\expandafter\ifx\csname urlstyle\endcsname\relax
  \providecommand{\doi}[1]{doi: #1}\else
  \providecommand{\doi}{doi: \begingroup \urlstyle{rm}\Url}\fi

\bibitem[Antoniou et~al.(2020)Antoniou, Patacchiola, Ochal, and
  Storkey]{Antoniou2020cfsl}
Antreas Antoniou, Massimiliano Patacchiola, Mateusz Ochal, and Amos Storkey.
\newblock {Defining Benchmarks for Continual Few-Shot Learning}.
\newblock \emph{arXiv preprint arXiv:2004.11967}, 2020.

\bibitem[Buda et~al.(2018)Buda, Maki, and Mazurowski]{Buda2018imbalance}
Mateusz Buda, Atsuto Maki, and Maciej~A. Mazurowski.
\newblock {A systematic study of the class imbalance problem in convolutional
  neural networks}.
\newblock \emph{Neural Networks}, 106, 2018.

\bibitem[Chen et~al.(2019)Chen, Wang, Liu, Kira, and Huang]{Chen2019closer}
Wei~Yu Chen, Yu~Chiang~Frank Wang, Yen~Cheng Liu, Zsolt Kira, and Jia~Bin
  Huang.
\newblock {A closer look at few-shot classification}.
\newblock \emph{International Conference on Learning Representations (ICLR)},
  2019.

\bibitem[Chen et~al.(2020)Chen, Dai, Li, Gao, and Song]{Chen2020mamlstop}
Xinshi Chen, Hanjun Dai, Yu~Li, Xin Gao, and Le~Song.
\newblock {Learning to Stop While Learning to Predict}.
\newblock \emph{International Conference on Machine Learning (ICML)}, 2020.

\bibitem[Dhillon et~al.(2020)Dhillon, Chaudhari, Ravichandran, and
  Soatto]{Dhillon2020baseline}
Guneet~Singh Dhillon, Pratik Chaudhari, Avinash Ravichandran, and Stefano
  Soatto.
\newblock {A Baseline for Few-Shot Image Classification}.
\newblock In \emph{International Conference on Learning Representations}, 2020.

\bibitem[Edwards \& Storkey(2017)Edwards and Storkey]{Edwards2017}
Harrison Edwards and Amos Storkey.
\newblock {Towards a Neural Statistician}.
\newblock \emph{International Conference on Learning Representations (ICLR)},
  2017.

\bibitem[Finn et~al.(2017)Finn, Abbeel, and Levine]{Finn2017maml}
Chelsea Finn, Pieter Abbeel, and Sergey Levine.
\newblock {Model-Agnostic Meta-Learning for Fast Adaptation of Deep Networks}.
\newblock \emph{International Conference on Machine Learning (ICML)}, 2017.

\bibitem[Guan et~al.(2020)Guan, Liu, Sun, Feng, Shuai, and
  Wang]{Guan2020aerial}
Jian Guan, Jiabei Liu, Jianguo Sun, Pengming Feng, Tong Shuai, and Wenwu Wang.
\newblock {Meta Metric Learning for Highly Imbalanced Aerial Scene
  Classification}.
\newblock In \emph{ICASSP 2020 - 2020 IEEE International Conference on
  Acoustics, Speech and Signal Processing (ICASSP)}. IEEE, 2020.

\bibitem[Hospedales et~al.(2020)Hospedales, Antoniou, Micaelli, and
  Storkey]{Hospedales2020}
Timothy Hospedales, Antreas Antoniou, Paul Micaelli, and Amos Storkey.
\newblock {Meta-Learning in Neural Networks: A Survey}.
\newblock \emph{arXiv preprint arXiv:2004.05439}, 2020.

\bibitem[Ioffe \& Szegedy(2015)Ioffe and Szegedy]{Ioffe2015batchnorm}
Sergey Ioffe and Christian Szegedy.
\newblock {Batch Normalization: Accelerating Deep Network Training by Reducing
  Internal Covariate Shift}.
\newblock \emph{International Conference on Machine Learning (ICML)}, 2015.

\bibitem[Johnson \& Khoshgoftaar(2019)Johnson and
  Khoshgoftaar]{Johnson2019imbalance}
Justin~M. Johnson and Taghi~M. Khoshgoftaar.
\newblock {Survey on deep learning with class imbalance}.
\newblock \emph{Journal of Big Data}, 6, 2019.

\bibitem[Lee et~al.(2019)Lee, Lee, Na, Kim, Park, Yang, and
  Hwang]{Lee2020baysiantaml}
Hae~Beom Lee, Hayeon Lee, Donghyun Na, Saehoon Kim, Minseop Park, Eunho Yang,
  and Sung~Ju Hwang.
\newblock {Learning to Balance: Bayesian Meta-Learning for Imbalanced and
  Out-of-distribution Tasks}.
\newblock \emph{International Conference on Machine Learning (ICML)}, 2019.

\bibitem[Leevy et~al.(2018)Leevy, Khoshgoftaar, Bauder, and
  Seliya]{Leevy2018bigdata}
Joffrey~L. Leevy, Taghi~M. Khoshgoftaar, Richard~A. Bauder, and Naeem Seliya.
\newblock {A survey on addressing high-class imbalance in big data}.
\newblock \emph{Journal of Big Data}, 5, 2018.

\bibitem[Liu et~al.(2019)Liu, Miao, Zhan, Wang, Gong, and Yu]{Liu2019tailed}
Ziwei Liu, Zhongqi Miao, Xiaohang Zhan, Jiayun Wang, Boqing Gong, and Stella~X.
  Yu.
\newblock {Large-Scale Long-Tailed Recognition in an Open World}.
\newblock \emph{IEEE Conference on Computer Vision and Pattern Recognition
  (CVPR)}, 2019.

\bibitem[Massiceti et~al.(2021)Massiceti, Theodorou, Zintgraf, Harris, Stumpf,
  Morrison, Cutrell, and Hoffmann]{massiceti2021orbit}
Daniela Massiceti, Lida Theodorou, Luisa Zintgraf, Matthew~Tobias Harris,
  Simone Stumpf, Cecily Morrison, Edward Cutrell, and Katja Hoffmann.
\newblock Orbit: A real-world few-shot dataset for teachable object recognition
  collected from people who are blind or low vision.
\newblock \emph{arXiv preprint arXiv:2104.03841}, 2021.

\bibitem[Ochal et~al.(2020)Ochal, Vazquez, Petillot, and Wang]{Ochal2020oceans}
Mateusz Ochal, Jose Vazquez, Yvan Petillot, and Sen Wang.
\newblock {A Comparison of Few-Shot Learning Methods for Underwater Optical and
  Sonar Image Classification}.
\newblock \emph{OCEANS 2020 preprint}, 2020.

\bibitem[Pan \& Yang(2010)Pan and Yang]{Pan2010transfer}
Sinno~Jialin Pan and Qiang Yang.
\newblock {A Survey on Transfer Learning}.
\newblock \emph{IEEE Transactions on Knowledge and Data Engineering}, 22, 2010.

\bibitem[Patacchiola et~al.(2020)Patacchiola, Turner, Crowley, O'Boyle, and
  Storkey]{Patacchiola2019gpshot}
Massimiliano Patacchiola, Jack Turner, Elliot~J. Crowley, Michael O'Boyle, and
  Amos Storkey.
\newblock {Bayesian Meta-Learning in the Few-Shot Setting via Deep Kernels}.
\newblock In \emph{Advances in Neural Information Processing Systems
  (NeurIPS)}, 2020.

\bibitem[Ravi \& Larochelle(2016)Ravi and Larochelle]{Ravi2017}
Sachin Ravi and Hugo Larochelle.
\newblock {Optimization as a model for few-shot learning}.
\newblock \emph{International Conference on Learning Representations (ICLR)},
  2016.

\bibitem[Reed(2001)]{Reed2001powerlaw}
William~J Reed.
\newblock The pareto, zipf and other power laws.
\newblock \emph{Economics Letters}, 2001.

\bibitem[Russakovsky et~al.(2015)Russakovsky, Deng, Su, Krause, Satheesh, Ma,
  Huang, Karpathy, Khosla, Bernstein, Berg, and Fei-Fei]{Russakovsky2015ilsvrc}
Olga Russakovsky, Jia Deng, Hao Su, Jonathan Krause, Sanjeev Satheesh, Sean Ma,
  Zhiheng Huang, Andrej Karpathy, Aditya Khosla, Michael Bernstein,
  Alexander~C. Berg, and Li~Fei-Fei.
\newblock {ImageNet Large Scale Visual Recognition Challenge}.
\newblock \emph{International Journal of Computer Vision}, 115, 2015.

\bibitem[{Salakhutdinov} et~al.(2011){Salakhutdinov}, {Torralba}, and
  {Tenenbaum}]{Salakhutdinov2011longtail}
R.~{Salakhutdinov}, A.~{Torralba}, and J.~{Tenenbaum}.
\newblock Learning to share visual appearance for multiclass object detection.
\newblock \emph{IEEE Conference on Computer Vision and Pattern Recognition
  (CVPR)}, 2011.

\bibitem[Snell \& Zemel(2020)Snell and Zemel]{Snell2020polya}
Jake Snell and Richard Zemel.
\newblock {Bayesian Few-Shot Classification with One-vs-Each Polya-Gamma
  Augmented Gaussian Processes}.
\newblock \emph{arXiv preprint arXiv:2007.10417}, 2020.

\bibitem[Snell et~al.(2017)Snell, Swersky, and Zemel]{Snell2017proto}
Jake Snell, Kevin Swersky, and Richard~S. Zemel.
\newblock {Prototypical Networks for Few-shot Learning}.
\newblock \emph{Advances in Neural Information Processing Systems (NeurIPS)},
  2017.

\bibitem[Sung et~al.(2017)Sung, Yang, Zhang, Xiang, Torr, and
  Hospedales]{Sung2017relationnet}
Flood Sung, Yongxin Yang, Li~Zhang, Tao Xiang, Philip H.~S. Torr, and
  Timothy~M. Hospedales.
\newblock {Learning to Compare: Relation Network for Few-Shot Learning}.
\newblock \emph{IEEE Conference on Computer Vision and Pattern Recognition
  (CVPR)}, 2017.

\bibitem[Triantafillou et~al.(2020)Triantafillou, Zhu, Dumoulin, Lamblin, Evci,
  Xu, Goroshin, Gelada, Swersky, Manzagol, and
  Larochelle]{Triantafillou2019meta}
Eleni Triantafillou, Tyler Zhu, Vincent Dumoulin, Pascal Lamblin, Utku Evci,
  Kelvin Xu, Ross Goroshin, Carles Gelada, Kevin Swersky, Pierre-Antoine
  Manzagol, and Hugo Larochelle.
\newblock {Meta-Dataset: A Dataset of Datasets for Learning to Learn from Few
  Examples}.
\newblock \emph{International Conference on Learning Representations (ICLR)},
  2020.

\bibitem[Vinyals et~al.(2017)Vinyals, Blundell, Lillicrap, Kavukcuoglu, and
  Wierstra]{Vinyals2017matching}
Oriol Vinyals, Charles Blundell, Timothy Lillicrap, Koray Kavukcuoglu, and Daan
  Wierstra.
\newblock {Matching Networks for One Shot Learning}.
\newblock \emph{Advances in Neural Information Processing Systems (NeurIPS)},
  2017.

\bibitem[Vogelbaum et~al.(2020)Vogelbaum, Dangovski, Jing, and
  Solja{\v{c}}i{\'{c}}]{Vogelbaum2020protocontext}
Evan Vogelbaum, Rumen Dangovski, Li~Jing, and Marin Solja{\v{c}}i{\'{c}}.
\newblock {Contextualizing Enhances Gradient Based Meta Learning}.
\newblock \emph{arXiv preprint arXiv:2007.10143}, 2020.

\bibitem[Wah et~al.(2011)Wah, Branson, Welinder, Perona, and
  Belongie]{Wah2011cub}
C.~Wah, S~Branson, P~Welinder, P~Perona, and S~Belongie.
\newblock {The Caltech-UCSD Birds-200-2011 Dataset}.
\newblock In \emph{California Institute of Technology}, 2011.

\bibitem[Wang et~al.(2014)Wang, Minku, and Yao]{wang2014resampling}
Shuo Wang, Leandro~L Minku, and Xin Yao.
\newblock Resampling-based ensemble methods for online class imbalance
  learning.
\newblock \emph{IEEE Transactions on Knowledge and Data Engineering},
  27\penalty0 (5), 2014.

\bibitem[Wang et~al.(2019)Wang, Chao, Weinberger, and van~der
  Maaten]{Wang2019simpleshot}
Yan Wang, Wei-Lun Chao, Kilian~Q Weinberger, and Laurens van~der Maaten.
\newblock {SimpleShot: Revisiting Nearest-Neighbor Classification for Few-Shot
  Learning}.
\newblock \emph{arXiv preprint arXiv:1911.04623}, 2019.

\bibitem[Wertheimer \& Hariharan(2019)Wertheimer and
  Hariharan]{Wertheimer2019metainat}
Davis Wertheimer and Bharath Hariharan.
\newblock {Few-Shot Learning with Localization in Realistic Settings}.
\newblock \emph{arXiv preprint arXiv:1904.08502}, 2019.

\bibitem[Zhang et~al.(2021)Zhang, Meng, Gouk, and Hospedales]{zhang2021shallow}
Xueting Zhang, Debin Meng, Henry Gouk, and Timothy Hospedales.
\newblock Shallow bayesian meta learning for real-world few-shot recognition.
\newblock \emph{arXiv preprint arXiv:2101.02833}, 2021.

\end{thebibliography}
\bibliographystyle{iclr2021_conference}

\clearpage
\appendix
\section{Implementation Details}\label{sec: implement}

\subsection{FSL Methods and Baselines}
In our experiments, we examined a wide range of meta-learning and few-shot learning models and baselines. Hyper-parameters are located in our source code.
\begin{enumerate}
    \item \textbf{Baseline (fine-tune)} \citep{Pan2010transfer,Chen2019closer} represents a classical way of applying transfer learning, where a neural network is pre-trained on a large dataset, then fine-tuned on a smaller domain-specific dataset. The backbone of the Baseline has a single linear classification layer with an output for each meta-training class. The network was trained during pre-training. During meta-testing, the pre-trained linear layer was exchanged for another randomly initialized layer with outputs matching the number of classes in the tasks ($N$-way). Fine-tuning was performed on the new randomly initialized classification layer using the support set $\mathcal{S}$.
    \item \textbf{Baseline\texttt{++}} \citep{Chen2019closer} augments the fine-tune baseline by using Cosine Similarity on the last layer.
    \item \textbf{Matching Network (Matching Net)} \citep{Vinyals2017matching} uses context embeddings with an LSTM to effectively perform k-nearest neighbor in embedding space using cosine similarity to classify the query set.
    \item \textbf{Prototypical Networks (ProtoNet)} \citep{Snell2017proto} maps images into a feature space and calculates class means (called prototypes). The query samples are then classified based on the closest Euclidian distance to class prototypes. We evaluate two models, the first meta-trained like the others on 5-way episodes, and the second trained on 20-way episodes. During 20-way meta-training, we set the query size to 5.
    \item \textbf{Relation Networks (Relation Net)} \citep{Sung2017relationnet} augment the classical Prototypical Networks by introducing a relation module (another neural network) that compares the distance using a learnable score. The original method uses Mean Squared Error to minimize the relation score between samples of the same type. However, we follow \cite{Chen2019closer} and use cross-entropy loss to improve performances. The structure of the relation module is described in section~\ref{sec backbone}. 
    \item \textbf{DKT} (formally called GPShot) proposed by \cite{Patacchiola2019gpshot} is a probabilistic approach that utilizes the Gaussian Processes with a deep neural network as a kernel function. We used Batch Norm Cosine distance for the kernel type. 
    \item \textbf{SimpleShot} \citep{Wang2019simpleshot} augments the 1-NN baseline model by normalizing and centering the feature vector using the mean feature vector of the dataset. The query samples are assigned to the nearest prototype according to the Euclidian distance. In contrast to the baseline models, pre-training is performed on the meta-training dataset like other meta-learning algorithms, and meta-validation is used to select the best model based on tasks sampled from $\mathcal{D}_{val}$. 
    \item \textbf{MAML} \citep{Finn2017maml} is a meta-learning technique that learns a common initialization of weights that can be quickly adapted for various tasks using fine-tuning on the support set. The task adaptation uses a standard gradient descent algorithm minimizing Cross-Entropy loss on the support set. The original method uses second-order derivates; however, due to more efficient calculation, we use the first-order MAML, which has been shown to work just as well. We set the inner-learning rate to 0.1 with 10 iteration steps based on our hyperparameter fine-tuning. We optimize the meta-learner model on batches of 4 meta-training tasks. 
    \item \textbf{ProtoMAML} \citep{Triantafillou2019meta} augments traditional first-order MAML by reinitializing the last classification layer between tasks. Specifically, the weights of the layer are assigned to the prototype for each corresponding output. This extra step combines the fine-tuning ability of MAML and the class regularization ability of Prototypical Networks. We set the inner-loop learning rate to 0.01 with 5 iterations. Unlike MAML, we found that updating the meta-learner after a single meta-training task gave the best performance.
\end{enumerate}

\subsection{Backbone Architectures}\label{sec backbone}
All methods shared the same backbone architecture. For the core contribution of our work, we used Conv4 architecture consisting of 4 convolutional layers with 64 channels (padding 1), interleaved by batch normalization \citep{Ioffe2015batchnorm}, ReLU activation function, and max-pooling (kernel size 2, and stride 2) \citep{Chen2019closer}. Relation Network used max-pooling only for the last 2 layers of the backbone to account for the relation module. The relation module consisted of two additional convolutional layers, each followed by batch norm, ReLU, and max-pooling.

\subsection{Datasets and Distributions}
We meta-trained methods on a variety of meta-training datasets and distributions. Specifically, we used Mini-ImageNet \citep{Ravi2017,Vinyals2017matching}, CUB-2011 \citep{Wah2011cub}, and ImageNet \cite{Russakovsky2015ilsvrc}. Our main set of experiments controls imbalance of the meta-training set of Mini-ImageNet \citep{Ravi2017} ($\mathcal{D}_{train}$). To estimate how significant dataset imbalance can be, we have to eliminate other factors that could influence the meta-learning performance. For this reason, and to have enough samples for the majority classes, we halve the total number of samples from the original $\mathcal{D}_{train}$ of Mini-ImageNet. We denote this halved dataset as $\mathcal{D}'_{train}$, where $|\mathcal{D}'_{train}| \approx 64 \times 300 = 19200$. We induce imbalance into the dataset according to one of the $\mathcal{I}$-distributions as outlined in Section~\ref{sec methodology}. For simplicity, we do not modify the distribution in the meta-validation and meta-testing datasets ($\mathcal{D}_{val}$ and $\mathcal{D}_{test}$), and we keep them the same as in the original Min-ImageNet. To emulate a stronger domain-shift scenario, we also evaluate trained models on 50 randomly selected classes from CUB-2011 \citep{Wah2011cub}. In Appendix~\ref{appendix reduced} we explore reducing the size of the meta-training dataset to contain $|\mathcal{D}''_{train}| \approx 32 \times 150 = 4800$. In Appendix~\ref{appendix longtail}, we induce long-tail imbalance in ImageNet \citep{Russakovsky2015ilsvrc}. More details can be found in the appropriate sections. 

\subsection{Meta-Training / Pre-Training}
All methods follow a similar three-phase learning procedure: meta-training, meta-validation, and meta-testing. During meta-training, an FSL model was exposed to 50k tasks sampled from $\mathcal{D}_{train}$ or imbalanced variants. After every 500 tasks, the model was validated on tasks from $\mathcal{D}_{val}$ and the best performing model was updated. At the end of the meta-training phase, the best model was evaluated on tasks sampled from $\mathcal{D}_{test}$. SimpleShot follows a similar three-phase procedure but with the meta-training phases exchanged for conventional pre-training on mini-batches of size 128. In all three meta-phases, we used 15 query samples per class, except for the 20-way Prototypical Network, where we used 5 query samples per class during meta-training to allow for a higher number of samples in the support set. All methods were meta-validated on 200 tasks/mini-batches every 250 meta-training tasks/mini-batches to select the best performing model. We used a learning rate of $1\times10^{-3}$ for the first 12.5k tasks, then reduced it to $1\times10^{-4}$ for the remaining tasks. 

\paragraph{Data Augmentation.} During the meta-/pre-training phases, we apply standard data augmentation techniques, following a similar setup to \cite{Chen2019closer}, with a random rotation of 10 degrees, scaling, random color/contrast/brightness jitter. Meta-validation and meta-testing had no augmentation. All images are resized to 84 by 84 pixels.

\subsection{Meta-Testing} The final test performances were measured on a random sample of 600 tasks. For all evaluations, we used standard, $balanced$ 5-shot 5-way tasks. We report the average one standard deviation in brackets and error-bars.

\newpage

\section{Verification of Implementation}\label{sec: verification}
We implement the FSL methods in PyTorch, adapting the implementation of \citep{Chen2019closer} but also borrowing from other implementations online (see individual method files in the source code for individual attribution). However, we have heavily modified these implementations to fit our imbalanced FSL framework, which also offers standard and continual FSL compatibility \citep{Antoniou2020cfsl}. We provide our implementations for ProtoMAML for which no open-source implementation in PyTorch existed. To verify our implementations, we compare methods on the standard balanced 5-shot 5-way task with reported accuracy. Results are presented in Table~\ref{tbl 5shot_5way}. We see that algorithms achieve very similar performance with no less than 3\% accuracy points compared to the reported performance. 

\begin{table}[htb]
    \caption[5-Shot 5-Way Classification]{Results of standard 5-shot 5-way experiments on Mini-ImageNet as achieved with our implementation compared to the original (reported) accuracy and other work. Other Sources's Accuracies were taken from: \ssymbol{1}~\citep{Chen2019closer}, \ssymbol{2}~\citep{Snell2020polya}, \ssymbol{3}~\citep{Vogelbaum2020protocontext}}\label{tbl 5shot_5way}
    \centering
    \scalebox{0.8}{
    \begin{tabular}{l|ccc}
\toprule
{} &                                       Acc (95\%CI) &                \specialcell{Original\\Acc(95\%CI)} &                      \specialcell{Other Sources\\Acc(95\%CI)} \\
Model                &                                                   &                                                   &                                                               \\
\midrule
Baseline (fine-tune) \citep{Chen2019closer} &           62.67\textcolor{gray}{\tiny{$\pm$0.70}} &           62.53\textcolor{gray}{\tiny{$\pm$0.69}} &                                                             - \\
Baseline++      \citep{Chen2019closer}     &  \textbf{66.43\textcolor{gray}{\tiny{$\pm$0.66}}} &           66.43\textcolor{gray}{\tiny{$\pm$0.63}} &                                                             - \\
Matching Net    \citep{Vinyals2017matching}     &           62.27\textcolor{gray}{\tiny{$\pm$0.69}} &           55.31\textcolor{gray}{\tiny{$\pm$0.73}} &           63.48\textcolor{gray}{\tiny{$\pm$0.66}} \ssymbol{1} \\
ProtoNet     \citep{Snell2017proto}        &           64.37\textcolor{gray}{\tiny{$\pm$0.71}} &           65.77\textcolor{gray}{\tiny{$\pm$0.70}} &           64.24\textcolor{gray}{\tiny{$\pm$0.72}} \ssymbol{1} \\
ProtoNet (20-way) \citep{Snell2017proto}   &           65.76\textcolor{gray}{\tiny{$\pm$0.70}} &  \textbf{68.20\textcolor{gray}{\tiny{$\pm$0.66}}} &  \textbf{66.68\textcolor{gray}{\tiny{$\pm$0.68}} \ssymbol{1}} \\
Relation Net (CE) \citep{Sung2017relationnet}   &           64.76\textcolor{gray}{\tiny{$\pm$0.68}} &           65.32\textcolor{gray}{\tiny{$\pm$0.70}} &           66.60\textcolor{gray}{\tiny{$\pm$0.69}} \ssymbol{1} \\
\bottomrule
\end{tabular}
    }
\end{table}

\clearpage

\section{Additional Experiments.}

\subsection{Small and Imbalanced Meta-Training Dataset}\label{appendix reduced}
In this section, we examine whether the effect of imbalance could be influenced by a smaller dataset size. Specifically, we construct a new set of datasets denoted by $\mathcal{D}''_{train}$ containing 1/8\textsuperscript{th} of samples in $\mathcal{D}_{train}$ of Mini-ImageNet, and 32 classes selected uniformly at random, $|\mathcal{D}''_{train}|=4800$. Table~\ref{tbl redu_mini_table_combined} are a break-down of Figure~\ref{fig reduced_meta}. Overall, the performance drops as the number of minority classes increases, but the effect still remains quite low ($<3\%$ absolute difference).

\begin{table}[htb]
    \caption{Evaluation accuracy after meta-training on $\mathcal{D}''_{train}$ derived from Mini-ImageNet with various imbalance distributions. Small differences in accuracy between $balanced$ and other distributions, suggest a small effect of imbalance at dataset level. \emph{Left:} Evaluation on the meta-testing dataset of Mini-ImageNet. \emph{Right:} Evaluation on the meta-testing dataset of CUB.
    } \label{tbl redu_mini_table_combined}
    \centering
    \scalebox{0.69}{ \begin{tabular}{r|cccc|cccc}
\toprule
$\mathcal{D}''_{train} \rightarrow \mathcal{D}_{test}$ & \multicolumn{4}{c|}{Mini-ImageNet $\rightarrow$ Mini-ImageNet} & \multicolumn{4}{c}{Mini-ImageNet $\rightarrow$ CUB} \\
Imbalance in $\mathcal{D}''_{train}$  &  \specialcell{ $balanced$ \\ {\scriptsize(150, 150, 32, -) }} & \specialcell{  $step$-8 \\ {\scriptsize(30, 190, 32, 8)} } & \specialcell{  $step$-16 \\ {\scriptsize(30, 270, 32, 16)} } & \specialcell{  $step$-24 \\ {\scriptsize(30, 510, 32, 24)} }  &  \specialcell{ $balanced$ \\ {\scriptsize(150, 150, 32, -) }} & \specialcell{  $step$-8 \\ {\scriptsize(30, 190, 32, 8)} } & \specialcell{  $step$-16 \\ {\scriptsize(30, 270, 32, 16)} } & \specialcell{  $step$-24 \\ {\scriptsize(30, 510, 32, 24)} } \\
\midrule
Baseline (fine-tune) &           56.07\textcolor{gray}{\tiny{$\pm$0.71}} &           56.19\textcolor{gray}{\tiny{$\pm$0.72}} &           56.17\textcolor{gray}{\tiny{$\pm$0.71}} &           55.81\textcolor{gray}{\tiny{$\pm$0.72}} &           55.80\textcolor{gray}{\tiny{$\pm$0.68}} &           55.19\textcolor{gray}{\tiny{$\pm$0.70}} &           55.14\textcolor{gray}{\tiny{$\pm$0.70}} &           53.38\textcolor{gray}{\tiny{$\pm$0.69}} \\
Baseline++           &           54.35\textcolor{gray}{\tiny{$\pm$0.67}} &           54.24\textcolor{gray}{\tiny{$\pm$0.67}} &           53.90\textcolor{gray}{\tiny{$\pm$0.67}} &           53.25\textcolor{gray}{\tiny{$\pm$0.68}} &           50.72\textcolor{gray}{\tiny{$\pm$0.69}} &           50.64\textcolor{gray}{\tiny{$\pm$0.70}} &           50.44\textcolor{gray}{\tiny{$\pm$0.69}} &           49.50\textcolor{gray}{\tiny{$\pm$0.67}} \\
\bottomrule
\end{tabular}} 
\end{table}

\clearpage

\subsection{Long-Tail Imbalance}\label{appendix longtail}
In this section, we explore the performance under a larger imbalance setting with long-tail distribution \citep{Salakhutdinov2011longtail}. We induce a long-tail distribution on classes from ImageNet \citep{Russakovsky2015ilsvrc}. We used ResNet-10 as the backbone. We partitioned ImageNet to contain 900 classes for $\mathcal{D}_{train}$ distributed according to $balanced$ or $\longtail$, while $\mathcal{D}_{val}$ and $\mathcal{D}_{test}$ contained 50 classes each with 500 randomly selected samples per class. The set of classes in each dataset is kept the same for all experiment repeats. We induce $\longtail$ imbalance in $\mathcal{D}_{train}$ using the Power-Law distribution \citep{Reed2001powerlaw} with a power of 10, a minimum of 20 samples per class (to allow 5-shot 15-query task), and a maximum of 1300 samples per class possible for ImageNet. This means that $\rho=65$ and the top 20\% majority classes in distribution account for 80\% of all data points in $\mathcal{D}_{train}$. We shuffled the classes within $\mathcal{D}_{train}$ such that the number of samples for a particular class could vary between repeats. The Balanced dataset contained samples distributed uniformly among the 900 classes (i.e. 137 samples per class) such that $|\mathcal{D}_{train}|\approx123300$ for both distributions (within 500 samples difference between them). Table~\ref{tbl longtail} shows the accuracy performance after training on 1800 balanced tasks across 3 seeds. The results show that meta-learner models -- ProtoNet, MAML, ProtoMAML-- are particularly susceptible to the larger imbalance, experiencing -6.3\%, -4.1\%, -4.5\% drop in accuracy, respectively. Interestingly, the Baseline seems to be the least affected by the imbalance with a non-significant drop in performance (-0.1\%). \hfill\vspace{0.5cm}
\begin{minipage}{0.5\linewidth}
  \centering
    \includegraphics[width=0.99\linewidth]{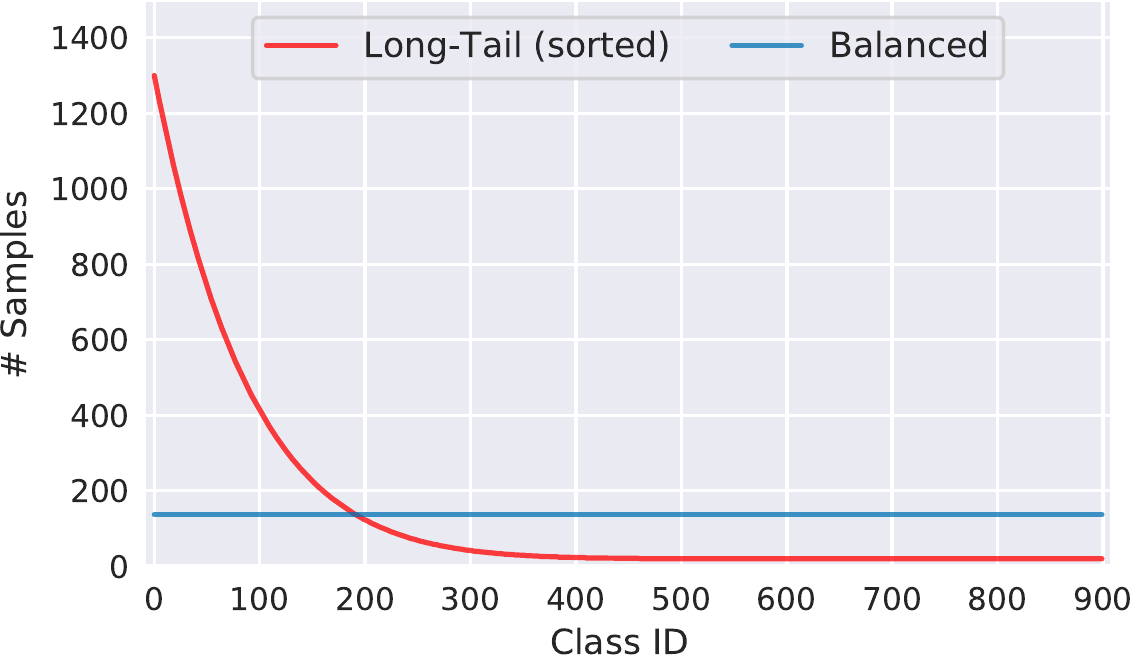}
    \captionof{figure}{Figure showing the class distribution for Balanced vs Long-Tail (sorted by class size).}
    \label{fig longtail}
\end{minipage}\hfill
\begin{minipage}{0.48\linewidth}
   \centering
    \captionof{table}{Accuracy performance on Long-Tail distribution vs balanced with approximately the same number of samples in total.} 
    \label{tbl longtail}
    \centering
    \scalebox{0.99}{
        \begin{tabular}{l|cc}
\toprule
$I$ in $\mathcal{D}_{train}$ & \specialcell{$balanced$\\{\scriptsize(137, 137, 900,-)}} & \specialcell{$\longtail$\\{\scriptsize(20, 1300, 900,-)}} \\
\midrule
Baseline & 62.57\textcolor{gray}{\tiny{$\pm$0.85}} & 62.47\textcolor{gray}{\tiny{$\pm$0.85}} \\
MAML     & 62.83\textcolor{gray}{\tiny{$\pm$0.90}} & 58.69\textcolor{gray}{\tiny{$\pm$0.90}} \\
ProtoNet & 67.97\textcolor{gray}{\tiny{$\pm$0.83}} & 61.67\textcolor{gray}{\tiny{$\pm$0.86}} \\
ProtoMAML & 65.44\textcolor{gray}{\tiny{$\pm$0.88}} & 60.97\textcolor{gray}{\tiny{$\pm$0.89}} \\
\bottomrule
\end{tabular}
    }
\end{minipage}

\end{document}